\documentclass[10pt,twocolumn,letterpaper]{article}

\usepackage{icb}
\usepackage{times}
\usepackage{epsfig}
\usepackage{graphicx}
\usepackage{amsmath}
\usepackage{amssymb}
\usepackage{subcaption}
\usepackage{booktabs}
\usepackage{tabularx}
\usepackage{caption}
\usepackage{subcaption}
\usepackage[perpage]{footmisc}

\usepackage{url}
\usepackage{hyperref}
\hypersetup{
    colorlinks
}

\DeclareMathOperator{\IR}{\mathbb{R}}
\DeclareMathOperator{\EL}{\mathcal{L}}
\DeclareMathOperator{\mcF}{\mathcal{F}}
\DeclareMathOperator{\mcG}{\mathcal{G}}
\DeclareMathOperator{\mcH}{\mathcal{H}}

\icbfinalcopy 

\ificbfinal\fi
\begin{document}

\title{ DocFace: Matching ID Document Photos to Selfies\thanks{Technically, the word ``selfies'' refer to self-captured photos from mobile phones. But here, we define ``selfies'' as any self-captured live face photos, including those from mobile phones and kiosks.} }

\author{Yichun Shi and Anil K. Jain\\
Michigan State University\\
East Lansing, Michigan, USA\\
{\tt\small shiyichu@msu.edu, jain@cse.msu.edu}
}

\maketitle
\thispagestyle{empty}

\begin{abstract}
   Numerous activities in our daily life, including transactions, access to services and transportation, require us to verify who we are by showing our ID documents containing face images, e.g. passports and driver licenses. An automatic system for matching ID document photos to live face images in real time with high accuracy would speed up the verification process and remove the burden on human operators. In this paper, by employing the transfer learning technique, we propose a new method, \textbf{DocFace}, to train a domain-specific network for ID document photo matching without a large dataset. Compared with the baseline of applying existing methods for general face recognition to this problem, our method achieves considerable improvement. A cross validation on an ID-Selfie dataset shows that DocFace improves the TAR from $61.14\%$ to $92.77\%$ at FAR=$0.1\%$. Experimental results also indicate that given more training data, a viable system for automatic ID document photo matching can be developed and deployed.
\end{abstract}

\section{Introduction}
\label{sec:intro}
\begin{figure}[t]
\centering
\includegraphics[width=\linewidth]{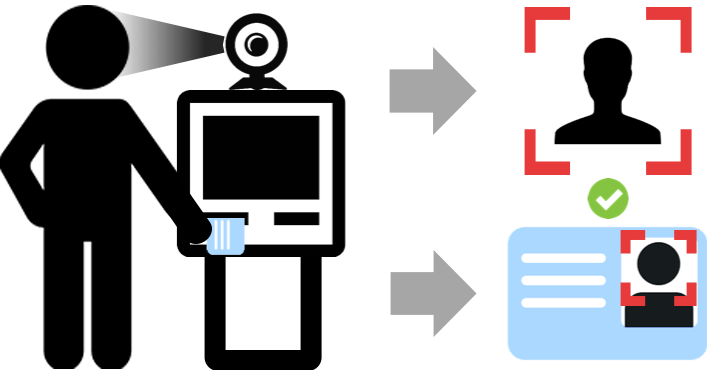}
\vspace{-1.5em}
\caption{An application scenario of the ID document matching system. The kiosk scans the ID document or reads its chip for the face photo and the camera takes another photo of the holder's live face (selfie). Then, through face recognition, the system decides whether the holder is indeed the owner of the ID document.}
\label{fig:application}
\vspace{-0.5em}
\end{figure}

Identity verification plays an important role in our daily lives. For example, access control, physical security and international border crossing require us to verify our access (security) level and our identities. Although an ideal solution is to have a digital biometric database of all citizens/users which can be accessed whenever user verification is needed, it is currently not feasible in most countries. A practical and common approach that has been deployed is to utilize an ID document containing the subject's photo, and verify the identity by comparing the document image with the subject's live face. For example, immigration and customs officials look at the passport photos and manually compare it with subject's face standing in front of them to confirm that the passenger is indeed the legitimate owner of the passport. Clerks at supermarkets look at driver licenses to check customers' age. Such an ID document photo matching is conducted in numerous scenarios, but it is primarily conducted by humans, which is time-consuming, costly and potentially error-prone. Therefore, an automatic system that matches ID document photos to selfies with high speed and low errors rate has a bright future in these applications. Here, we define the ``selifes'' as any self-captured photos, including those from kiosks. In addition, automatic ID matching systems also enable remote applications not otherwise possible, such as onboarding new customers in a mobile app (verifying their identities for account creation), or account recovery. One application scenario of ID document photo matching system is illustrated in Figure ~\ref{fig:application}.

\begin{figure}
\center
\begin{subfigure}[b]{\linewidth}
    \includegraphics[width=\linewidth]{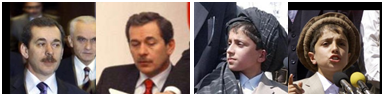}
    \caption{General face matching}
\end{subfigure}
\begin{subfigure}[b]{\linewidth}
    \includegraphics[width=\linewidth]{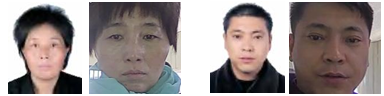}
    \caption{ID document photo matching}
\end{subfigure}
\caption{Example images from (a) LFW dataset~\cite{LFWTech} and (b) ID-Selfie-A dataset. Each row shows two pairs from the two datasets, respectively. Compared with the general unconstrained face recognition shown in (a), ID Document photo matching (b) does not need to consider large pose variations. Instead, it involves a number of other challenges such as aging and information loss via image compression.}
\label{fig:examples}
\end{figure}

\begin{figure*}
\center
\begin{subfigure}[b]{0.32\linewidth}
    \includegraphics[width=\linewidth]{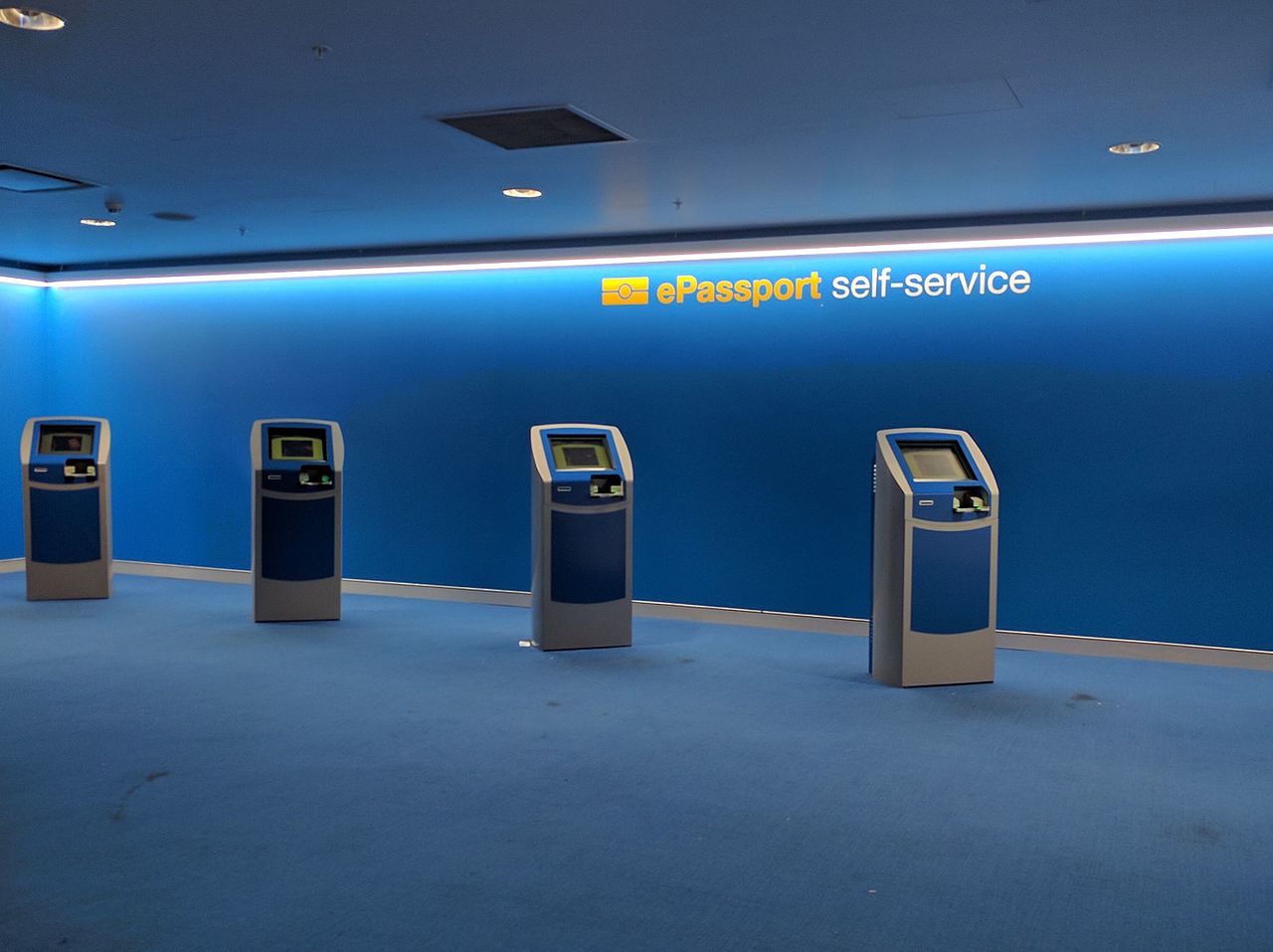}
    \caption{Australia SmartGate~\cite{gates_au}}
\end{subfigure}
\begin{subfigure}[b]{0.32\linewidth}
    \includegraphics[width=\linewidth]{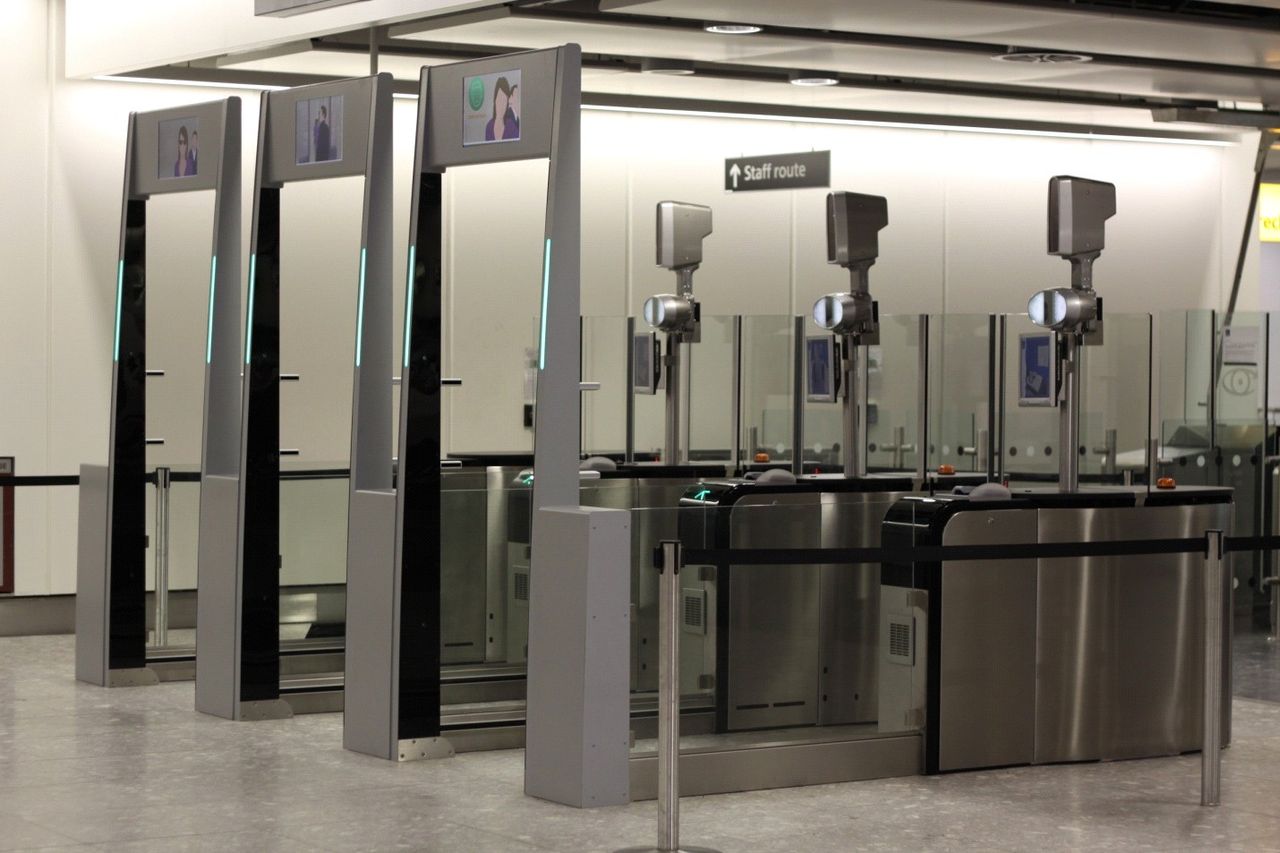}
    \caption{UK ePassport gates~\cite{gates_uk}}
\end{subfigure}
\begin{subfigure}[b]{0.32\linewidth}
    \includegraphics[width=\linewidth]{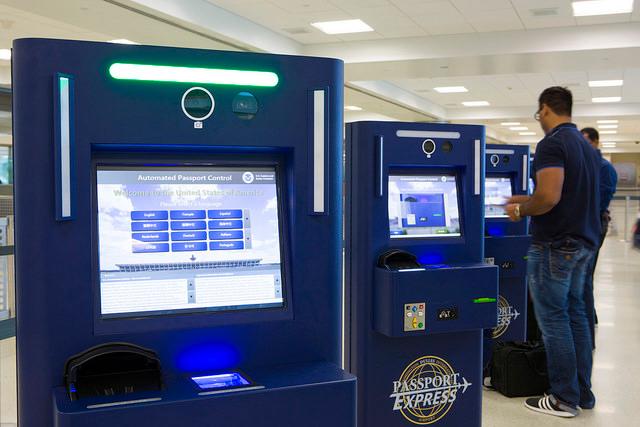}
    \caption{US Automated Passport Control~\cite{gates_us}}
\end{subfigure}
\caption{Example automatic ID document photo matching systems at international borders.}
\label{fig:gates}
\end{figure*}
A number of automatic ID document photo to selfies matching systems have been deployed at international borders. The earliest of such a system is the SmartGate deployed in Australia~\cite{gates_au}. See Figure~\ref{fig:gates}. Due to the increasing number of travelers to Australia, the Australian government introduced SmartGate at most of its international airports for electronic passport control checks for ePassport holders. To use the SmartGate, travelers only need to let a machine read their ePassport chips containing their digital photos and then capture their face images using a camera mounted at the SmartGate. After verifying a traveler's identity by face comparison, the gate is automatically opened for the traveler to enter Australia. Similar machines have also been installed in the UK (ePassport gates) ~\cite{gates_uk}, USA (US Automated Passport Control)~\cite{gates_us} and other countries. However, all of the above border crossing applications read the subject's face image from ePassport's chip. If a traveler does not have an ePassport, he will still have to be processed by an inspector who will do the typical manual photo comparison. Besides international border crossing, some businesses are providing face recognition solutions to ID document verification for online services~\cite{netverify}~\cite{mitek}.


The problem of ID document face matching involves many difficulties that are different from general face recognition, namely image to image matching. For typical unconstrained face recognition tasks, the main difficulties lie in the pose, illumination and expression (PIE) variations. But in document photo matching, we are comparing a scanned or digital document photo to a digital camera photo of a live face. Assuming that the user is cooperative, both of the images are captured under constrained conditions and there would not be large PIE variations. Instead, low quality of document photos due to image compression\footnote{Most chips in e-Passports have a memory from 8KB to 30KB; the face images have to be compressed to be stored in the chip. See \url{https://www.readid.com/blog/face-images-in-ePassports}} and the time gap between document issue date and verification time remain as the main difficulties, as shown in Figure~\ref{fig:examples}. In addition, since nowadays all of the face recognition systems use neural networks, another difficulty faced in our problem is the lack of a large dataset (pairs of ID photos and selfies), which is crucial to the training and evaluation of deep neural networks.

In spite of these numerous applications and challenges, there is paucity of research on this topic. On the one hand, only a few studies have been published on ID document matching\cite{starovoitov2000matching}\cite{bourlai2009matching}\cite{bourlai2011restoring}\cite{starovoitov2002three}, all of which are over five years old. On the other hand, face recognition technology has made tremendous strides in the past five years, mainly due to the availability of large scale face training data and deep neural network models for face recognition. Verification Rate (VR) on the Labeled Faces in the Wild (LFW) dataset, one of the first public domain "faces in the wild" dataset, at a False Accept Rate (FAR) of $0.1\%$ has increased from $41.66\%$ in 2014~\cite{liao2014benchmark} to $98.65\%$ in 2017~\cite{hasnat2017deepvisage}. Hence, the earlier published results on ID document photo to live face matching are now obsolete. Advances in face recognition algorithms allow us to build more robust and accurate matchers for ID document matching.

In this paper, we first briefly review existing studies on the ID document photo matching problem and state-of-the-art deep neural network-based face recognition methods. We then propose DocFace for developing a domain-specific face matcher for ID document photos by exploiting transfer learning techniques. Our experiments use two datasets of Chinese Identity Cards with corresponding camera photos to evaluate the performance of a Commercial-Off-The-Shelf (COTS) face matcher, open source deep network face matcher and the proposed method. The contributions of the paper are summarized below:
\begin{itemize}
    \item An evaluation of published face matchers on the problem of ID document photo matching.
    \item A new system and loss function for learning representations from heterogeneous face pairs.
    \item A domain-specific matcher, namely DocFace, for ID Document photo matching, which significantly improves the performance of existing general face matchers. The TAR on a private Chinese Identity Card dataset is improved from $61.14\%$ to $92.77\%$ at FAR=$0.1\%$.
\end{itemize}

\section{Related Works}

\subsection{ID Document Photo Matching}
To the best of our knowledge, the first study on ID document face photo matching is attributed to Starovoitov et al.~\cite{starovoitov2000matching}~\cite{starovoitov2002three}. Assuming all face images are frontal faces without large expression variations, the authors first localize the eyes with Hough Transform. Based on eye location, face region is cropped and gradient maps are computed as feature maps. The algorithm is similar to a general constrained face matcher, except it is developed for a document photo dataset. Bourlai et al. [13][14] considered ID document face recognition as a comparison between degraded face images by scanning the document photo against high quality live face images. Because their dataset is composed of scanned document photos (i.e. passports), they categorized the degradation of scanned document photos into three types: i) person-related, including hairstyle, makeup and aging; ii) document related, including image compression and watermarks; iii) scanning-device related, such as operator variability. To eliminate the degradation caused by these factors, Bourlai et al. inserted an image restoration phase before comparing the photos using a general face matcher. In particular, they train a classifier to classify the degradation type for a given image, and then apply a degradation-specific linear and nonlinear filters to restore the degraded images. Compared with their work on scanned documents, the document photos in our dataset are read from the chips in the Chinese Identity Cards. Additionally, our method is not designed for any specific degradation type but could be applied to any ID document photos.

\begin{figure*}[t]
\centering
\begin{subfigure}[b]{0.325\linewidth}
    \centering
    \includegraphics[width=0.9\linewidth]{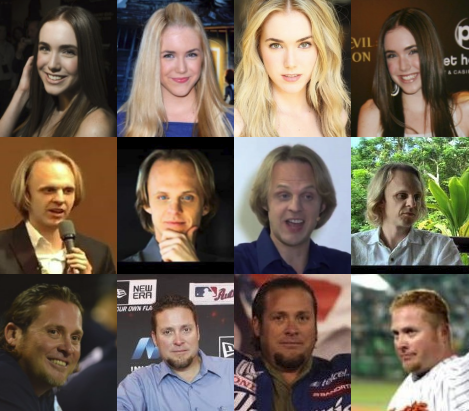}
    \caption{MS-Celeb-1M}
    \label{fig:datasets:ms}
\end{subfigure}
\begin{subfigure}[b]{0.325\linewidth}
    \centering
    \includegraphics[width=0.9\linewidth]{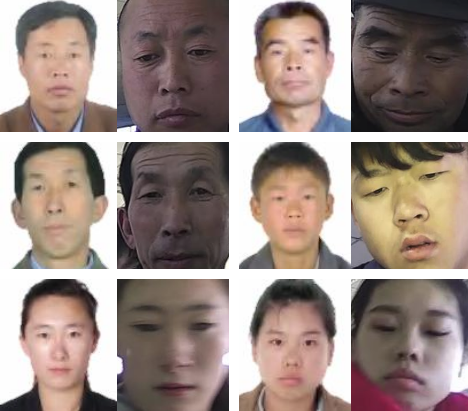}
    \caption{ID-Selfie-A}
    \label{fig:datasets:zk10k}
\end{subfigure}
\begin{subfigure}[b]{0.325\linewidth}
    \centering
    \includegraphics[width=0.9\linewidth]{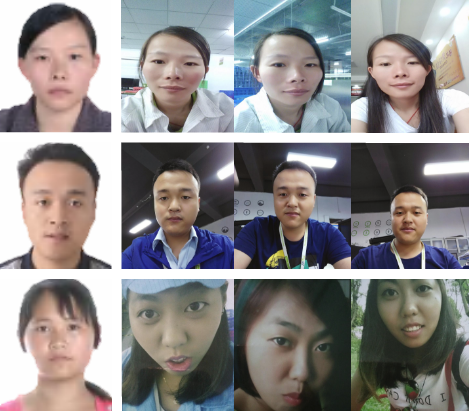}
    \caption{ID-Selfie-B}
    \label{fig:datasets:zk1n}
\end{subfigure}
\vspace{-0.5em}
\caption{Example images in each dataset. The left image in each pair in (b) and each row in (c) is the ID photo and on its right are the corresponding selfies.}
\label{fig:datasets}
\vspace{-1.0em}
\end{figure*}

\subsection{Deep Face Recognition}
Since the success of deep neural networks in the ImageNet competition~\cite{krizhevsky2012imagenet}, all of the ongoing research and development in face recognition now utilizes deep neural networks to learn face representations~\cite{taigman2014deepface}~\cite{deepid2plus}~\cite{schroff2015facenet}~\cite{hasnat2017deepvisage}. The popularity of deep neural networks could partially be attributed to a special property that the low-level image features are transferable, i.e. they are not limited to a particular task, but applicable to many image analysis tasks. Given this property, one can first train a network on a large dataset to learn salient low-level features, then train a domain specific neural network by transfer learning on a relatively small dataset. For example, Sankaranarayanan et al.~\cite{sankaranarayanan2016triplet} proposed to retrain networks by using a triplet probability embedding (TPE) loss function and achieved good results on the IJB-A benchmark~\cite{klare2015pushing}. Xiong et al.~\cite{xiong2017good} proposed a framework named Transferred Deep Feature Fusion (TDFF) to fuse the features from two different networks trained on different datasets and learn a face classifiers in the target domain, which achieved state-of-the-art performance on IJB-A dataset. Mittal et al.~\cite{mittal2015composite} developed a sketch-photo face matching system by fine-tuning the features of a stacked Auto-encoder and Deep Belief Network trained on a larger unconstrained face dataset.

\section{Datasets}
In this section we briefly introduce the datasets that are used in this paper. An overview of the datasets are in Figure~\ref{fig:datasets}.

\subsection{MS-Celeb-1M}
The MS-Celeb-1M dataset~\cite{guo2016msceleb} is a public domain face dataset facilitating training of deep networks for face recognition. It contains $8,456,240$ images of $99,892$ subjects (mostly celebrities) downloaded from internet. In our transfer learning framework, it is used as the source domain to train a very deep network with rich low-level features.
However, the dataset is known to have many mislabels. We use a cleaned version of MS-Celeb-1M with $5,041,527$ images of $98,687$. Some example images are shown in Figure~\ref{fig:datasets:ms}


\subsection{ID-Selfie-A Dataset}
\label{sec:dataset_A}
Our first ID document-selfie dataset is a private dataset composed of $10,000$ pairs of ID Cards photo and selfies. The ID card photos are read from chips in the Chinese Resident Identity Cards\footnote{\url{https://en.wikipedia.org/wiki/Resident_Identity_Card}}. The selfies are from a stationary camera. Among the $10,000$ pairs, we were able to align only $9,915$ pairs, i.e. a total of $19,830$ images. Assuming all the participants are cooperative, and hence there should be no failure-to-enroll case, we only keep these aligned pairs for our experiments. This dataset represents the target domain in our transfer learning framework. In experiments, we will further separate the dataset into two parts, one part to fine-tune the network trained on source domain, the other part for testing the performance. Some example pairs from this dataset are shown in Figure~\ref{fig:datasets:zk10k}.

\subsection{ID-Selfie-B Dataset}
Our second ID document-selfie dataset is a private dataset composed of $10,844$ images from $547$ subjects, each with one ID Card image and a varying number of selfies from different devices, including mobile phones. Compared with ID-Selfie-A, the selfies in this dataset are less constrained and some images have been warped or processed by image filters, as shown in Figure~\ref{fig:datasets:zk1n}. Out of these $547$ subjects, some subjects do not have any selfie photos. After cleaning and alignment, we retain $10,806$ images from $537$ subjects and use them for cross-dataset evaluation of the model trained on ID-Selfie-A dataset. There is no overlapping between the identities in ID-Selfie-A dataset and those in ID-Selfie-B dataset. See Figure~\ref{fig:datasets:zk1n} for example images in this dataset.

\section{Methodology}
\subsection{Notation}
\begin{figure}[t]
\centering
\includegraphics[width=\linewidth]{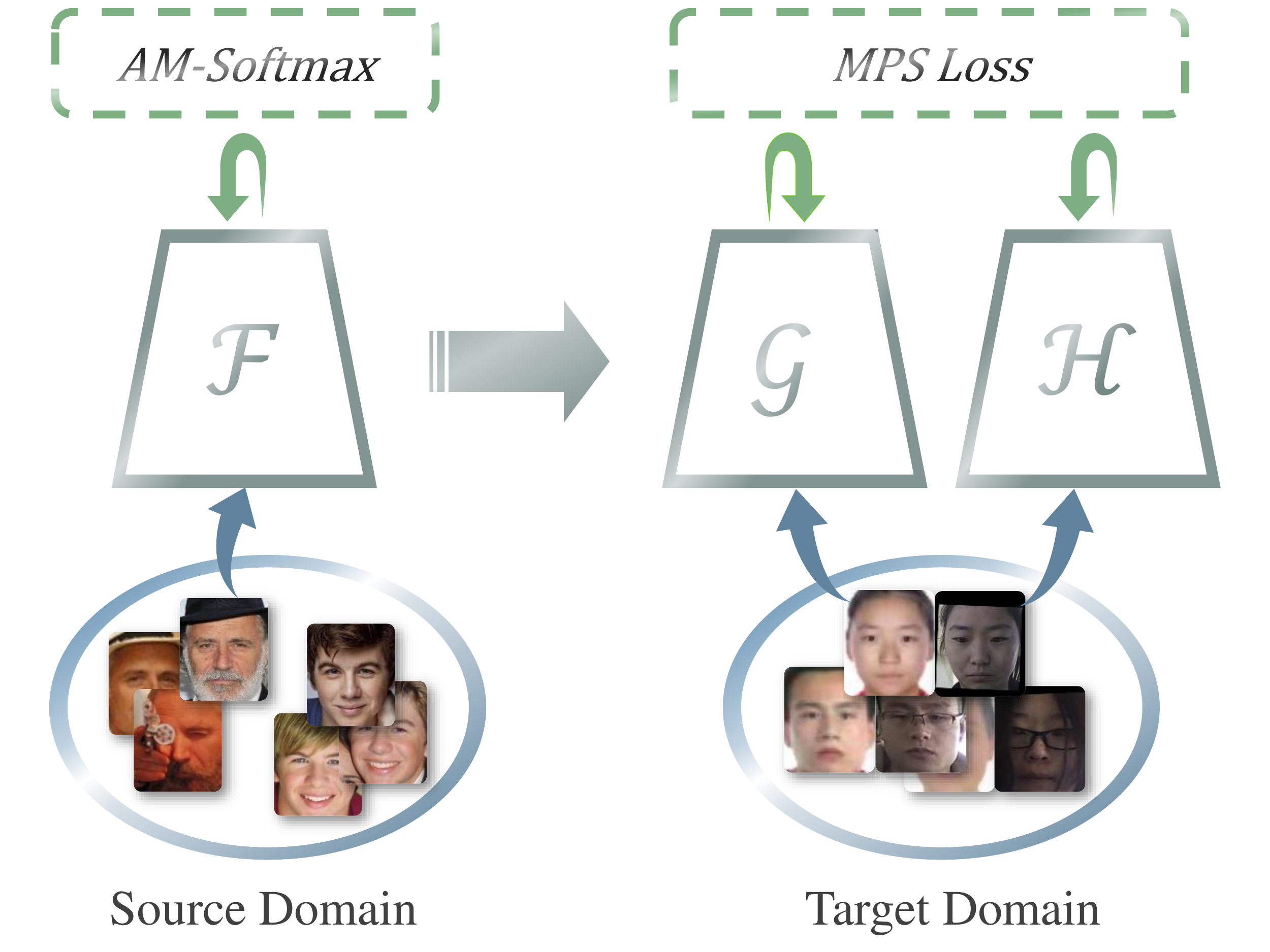}
\vspace{-1.5em}
\caption{Overview of the work flow of the proposed method. We first train a base model $\mcF$ on a large scale unconstrained face dataset. Then the features are transferred to domain-specific models $\mcG$ and $\mcH$, which are trained on a ID-Selfie dataset using the proposed MPS loss function.}
\label{fig:overview}
\vspace{-0.5em}
\end{figure}

Using a transfer learning framework, we first train a network as \emph{base model} on the source domain, i.e. unconstrained face dataset and then transfer its features to the target domain, ID and Selfie face images. Let $X^s=\{(x^s_i, y^s_i)|i=1,2,3,\cdots, N^s\}$ be the dataset of source domain, where $x^s_i\in\IR^{h\times w}$ and $y^s_i= {1,2,3,\cdots,C}$ are the $i^{th}$ image and label, respectively, $h$ and $w$ are the height and width of images, $N^s$ is the number of images, and $C$ is the number of classes. The training dataset of the target domain is denoted by $X^t=\{(x^t_{i1}, x^t_{i2})|i=1,2,3,\cdots, N^t\}$ where $x^t_{i1}\in\IR^{h\times w}$ and $x^t_{i2}\in\IR^{h\times w}$ refer to the ID image and selfie of the $i^{th}$ subject in the source domain, respectively. Here, $N^t$ is the number of ID-selfie pairs rather than the number of images. Function $\mcF:\IR^{h\times w} \rightarrow \IR^{d}$ denotes the base model for the source domain where $d$ is the dimensionality of the face representation. Similarly, $\mcG:\IR^{h\times w} \rightarrow \IR^{d}$ represents the face representation network for ID photos and $\mcH:\IR^{h\times w} \rightarrow \IR^{d}$ for selfies. An overview of the work flow is shown in Figure~\ref{fig:overview}.

\subsection{Training on source domain}
The source domain in our work is unconstrained face recognition, where we can train a very deep network on a large-scale dataset composed of different types of face images from a large number of subjects, i.e. MS-Celeb-1M. The objective is to train a base model $\mcF$ so that its face representations maximizes inter-subject separation and minimizes intra-subject variations. To guarantee its performance for better transfer learning, we utilize the popular Face-ResNet architecture~\cite{hasnat2017deepvisage} to build the convolutional neural network. We adopt the state-of-the-art \textit{Additive Max-margin Softmax} (AM-Softmax) loss function~\cite{wang2018additive}\cite{deng2018arcface}\cite{wang2018cosface} for training the base model. For each training sample in a mini-batch, the loss function is given by:

\begin{equation}
\newcommand{\softmax}{-\log\frac{ \exp(s\cos{\theta_{y_i,i}}-m) }{ \exp(s\cos{\theta_{y_i,i}}-m) + \sum_{j\neq y_i}{\exp(s\cos{\theta_{j,i}})}  }}
    \EL_s ={\softmax}
\label{eq:loss_additive}
\end{equation}
where
\begin{align*}
    \cos{\theta_{j,i}} = W_j^Tf_i \\
    W_j=\frac{W_j^*}{\|W_j^*\|^2_2} \\
    f_i=\frac{\mcF(x^s_i)}{\|\mcF(x^s_i)\|^2_2}.
\end{align*}
$W^*\in\IR^{d\times C}$ is the weight matrix and $m$ is a hyper-parameter for controlling the margin. The scale parameter $s$ can either be manually chosen or automatically learned~\cite{wang2017normface}; we leave it automatically learned for simplicity. During training, the loss in Equation~(\ref{eq:loss_additive}) is averaged across all images in the mini-batch.

\subsection{Training on target domain}
\label{sec:method_target}
The target domain is a relatively small dataset composed of ID-selfie image pairs. The sources of these images are very different from those from the source domain, thus directly applying $\mcF$ to these images will not work well. Because the ID images and selfies are from different sources, the problem can also be regarded as a sub-problem of the heterogeneous face recognition~\cite{klare2013heterogeneous}. A common approach in heterogeneous face recognition is to utilize two separate domain-specific models to map images from different sources into a unified embedding space. Therefore, we use a pair of \textit{sibling networks} $\mcG$ and $\mcH$ for ID images and selfie images, respectively, which share the same architecture but could have different parameters. Both of their features are transferred from $\mcF$, i.e. they have the same initialization. Notice that although this increases the model size, the inference speed would remain unchanged as each image is only fed into one of the sibling networks. 


Inspired by recent metric learning methods~\cite{schroff2015facenet}, we propose a \textit{Max-margin Pairwise Score} (MPS) loss for training the heterogeneous face pair dataset. For each mini-batch of size $M$, $M/2$ ID-selfie pairs are randomly selected from all the subjects. For each pair, the MPS loss is given by:

\begin{equation}
    \EL_t = [\max_{j\neq i}(\max(\cos{\theta_{j,i}},\cos{\theta_{i,j}})) - \cos{\theta_{i,i}} + m']_+
\label{eq:loss_hetero}
\end{equation}
where
\begin{align*}
    \cos{\theta_{i,j}} = g_i^Th_j \\
    g_i=\frac{\mcG(x^t_{i1})}{\|\mcG(x^t_{i1})\|^2_2} \\
    h_i=\frac{\mcH(x^t_{i2})}{\|\mcH(x^t_{i2})\|^2_2}.
\end{align*}
The loss is averaged across all the $M/2$ pairs. Here, $j$ iterates over all the other subjects in the batch. $[x]_+=max(0,x)$. Hyper-paramter $m'$  is similar to the $m$ in the AM-Softmax. The idea of MPS loss function in Equation~(\ref{eq:loss_hetero}) is learning representation by maximizing the margin between genuine pair similarities and imposter pair similarities. The MPS loss simulates the application scenario where the ID photos act like templates, while selfies from different subjects act like probes trying to be verified, or conversely. Notice that, after the hardest imposter pair is chosen by maximum score, the MPS loss is similar to Triplet Loss~\cite{schroff2015facenet} with one of the ID / selfie image as the anchor.

\section{Experiments}
\label{sec:exp}
\subsection{Experiment Settings}
\label{sec:exp_settings}
We conduct all of our experiments using Tensorflow r$1.2$. When training the base model on the MS-Celeb-1M, we use a batch size of $256$ and keep training for $280$K steps. We start with a learning rate of $0.1$ and it is decreased to $0.01$, $0.001$ after $160$K and $240$K steps, respectively. When fine-tuning on the ID-Selfie-A dataset, we keep the batch size of $256$ and train the sibling networks for $800$ steps. We start with a lower learning rate of $0.01$ and decrease the learning rate to $0.001$ after $500$ steps. For both the training stages, the model is optimized by Stochastic Gradient Descent (SGD) optimizer with a momemtum of $0.9$ and a weight decay of $(5e-4)$. All the images are aligned via similarity transformation based on landmarks detected by MTCNN~\cite{zhang2016joint} and resized to $96\times112$. We set margin parameters $m$ and $m'$ as $5.0$ and $0.5$, respectively. All the training and testing are run on a single Nvidia Geforce GTX 1080Ti GPU with $11$GB memory. The inference speed of our model on this GPU is $0.003$s per image.

By utilizing the MS-Celeb-V1 dataset and the AM-Softmax loss function in Equation~(\ref{eq:loss_additive}), our Face-ResNet network achieved $99.67\%$ accuracy on the standard verification protocol of LFW and a Verification Rate (VR) of $99.60\%$ at False Accept Rate (FAR) of $0.1\%$ on the BLUFR~\cite{liao2014benchmark} protocol. 

We name our method as DocFace. In Section~\ref{sec:exp_explore}, we first conduct a few exploratory experiments on the ID-Selfie-A dataset to compare different approaches for training a ID-Selfie matcher and to justify the efficacy of proposed methods. Then in Section~\ref{sec:exp_compare}, we compare the performance of DocFace with existing general face matchers on the ID-Selfie-A dataset. In Section~\ref{sec:exp_dataset}, by training the model on different subsets of the ID-Selfie-A dataset, we show that the performance increases steadily with the increase of dataset size in target domain. Finally, by using the model trained on ID-Selfie-A dataset, we conduct a cross-dataset evaluation on the ID-Selfie-B dataset.

For all the experiments on ID-Selfie-A dataset, a five-fold cross validation is conducted to evaluate the performance and robustness of the methods. The dataset is equally split into 5 splits, and in each fold one split is used for testing while the remaining are used for training. In particular, $7,932$ and $1,983$ pairs are used for training and testing, respectively, in each fold. We use the whole ID-Selfie-B dataset for cross-dataset evaluation. Cosine similarity is used as comparison score for all experiments.

\subsection{Exploratory Experiments}
\label{sec:exp_explore}

\begin{table}[t]
\scriptsize
\begin{center}
\begin{tabularx}{\linewidth}{ccccc}
\toprule
Model                   & Loss          &  Sibling         & VR(\%)            & VR(\%)  \\
                        &               &  Networks        & @FAR$=0.01\%$                  & @FAR$=0.1\%$  \\
\midrule                            
FS                      & MPS           &  Yes              & $0.03\pm0.04$                & $0.07\pm0.04$                 \\
BM                      & -             &  No              & $67.88\pm1.72$              & $82.06\pm1.40$                \\
TL                      & L2-Softmax    &  Yes              & $70.53\pm1.73$              & $85.15\pm1.38$                \\
TL                      & AM-Softmax    &  Yes              & $71.07\pm1.81$              & $85.24\pm1.52$                \\
TL                      & MPS           &  No              & $85.71\pm1.29$              & $92.51\pm1.13$                \\
TL                      & MPS           &  Yes              & $\mathbf{86.27\pm1.39}$   & $\mathbf{92.77\pm1.03}$         \\
\bottomrule
\end{tabularx}
\end{center}
\vspace{-1.5em}
\caption{\small Performance of different approaches for developing a ID Face matcher on the ID-Selfie-A dataset. ``FS'',``BM'' and ``TL'' refer to ``from scratch'', ``base model'' and ``transfer learning'', respectively. ``VR'' refers to Verification Rate. For the pre-trained model, because there is no training involved, we leave the loss function as blank.}
\label{tab:training_strategies}
\vspace{-1.5em}
\end{table}

In this section, by using the ID-Selfie-A dataset, we compare different ways to develop an ID-Selfie face matcher. First, we compare with the approaches without transfer learning: (1) a network trained from scratch with the same architecture and MPS loss function, and (2) the base model pre-trained on MS-Celeb-1M but not fine-tuned. To justify the efficacy of the proposed MPS loss function, we fine-tune the base model on ID-Selfie dataset using two other loss functions:  L2-Softmax~\cite{ranjan2017l2} and AM-Softmax~\cite{wang2018additive}, which achieved successful results in unconstrained face recognition. Finally, using the base model and MPS loss function, we compare the performance of sibling networks, i.e. different parameters for $\mcG$ and $\mcH$, to that of a shared network, i.e. $\mcG=\mcH$. As mentioned in Section~\ref{sec:exp_settings}, all the experiments are conducted in five fold cross validation and we report the average performance and standard deviation.

The results are shown in Table~\ref{tab:training_strategies}. Because the ID-Selfie-A dataset is too small, the model trained from scratch (FS) overfits heavily and performs very poorly. Similar result was observed even when we were trying to train a smaller network from scratch. In comparison, the base model (BM) pre-trained on MS-Celeb-V1 perform much better even before fine-tuning. This confirms that the features learned by the neural networks are transferable and can be helpful for developing domain specific matchers with small dataset. This performance is then further improved after transfer learning (TL). Although both L2-Softmax and AM-Softmax lead to an improvement in the performance, our proposed loss function (MPS) outperforms the pre-trained even more significantly. This is because our loss function is specially designed for the problem, and directly maximizes the margin of pairwise score rather than classification probability. Finally, we find that using a pair of sibling networks $\mcG$ and $\mcH$ slightly outperforms that of using a shared network. This means learning separate domain-specific models for ID photos and selfies could help the system learn more discriminative low-level features and lead to better face representations in the shared embedding space.


\begin{table}[t]
\scriptsize
\begin{center}
\begin{tabularx}{\linewidth}{Xccc}
\toprule
Method                              & \multicolumn{2}{c}{VR(\%) on ID-Selfie-A}   & VR(\%) on LFW \\
\cmidrule(lr){2-3}\cmidrule(lr){4-4}
                                    & @FAR$=0.01\%$        & @FAR$=0.1\%$          &  @FAR$=0.1\%$ \\      
\midrule
COTS                                & $27.32\pm1.46$      & $46.33\pm1.61$         & $92.01$           \\
CenterFace~\cite{wen2016discriminative}                 & $28.02\pm1.93$    & $60.10\pm1.68$        &  $91.70$   \\
SphereFace~\cite{liu2017sphereface}                     & $34.76\pm0.88$    & $61.14\pm0.82$        &  $96.74$   \\
\textit{DocFace}                       & $\mathbf{86.27\pm1.39}$   & $\mathbf{92.77\pm1.03}$        &  -       \\
\bottomrule
\end{tabularx}
\end{center}
\vspace{-1.5em}
\caption{\small Comparison of the proposed method with existing general face matchers on the ID-Selfie-A dataset under five-fold cross validation protocol. ``VR'' refers to Verification Rate. ``TL'' refers to Transfer Learning. For comparison, we report the performance of existing matchers on LFW according to BLUFR protocol~\cite{liao2014benchmark}. The proposed model is shown in italic style.}
\label{tab:existing_methods}
\vspace{-1.5em}
\end{table}

\subsection{Comparison with Existing Matchers}
\label{sec:exp_compare}
We evaluate the performance of existing general face matchers on the ID-Selfie-A dataset and compare them with the proposed method. To make sure our experiments are comprehensive enough, we compare our method not only with a Commercial-Off-The-Shelf (COTS) matcher, but also two open-source matchers representing the state-of-the-art unconstrained face recognition methods: CenterFace\footnote{\url{https://github.com/ydwen/caffe-face}}~\cite{wen2016discriminative} and SphereFace\footnote{\url{https://github.com/wy1iu/sphereface}}~\cite{liu2017sphereface}. During the five-fold cross validation, because the existing methods don't involve training, only the test split is used. For comparison, we also report the performance of the existing matchers on the unconstrained face dataset, LFW~\cite{LFWTech}, using the BLUFR protocol~\cite{liao2014benchmark}.

\begin{figure}[t]
\centering
\begin{subfigure}[b]{\linewidth}
    \centering
    \includegraphics[width=\linewidth]{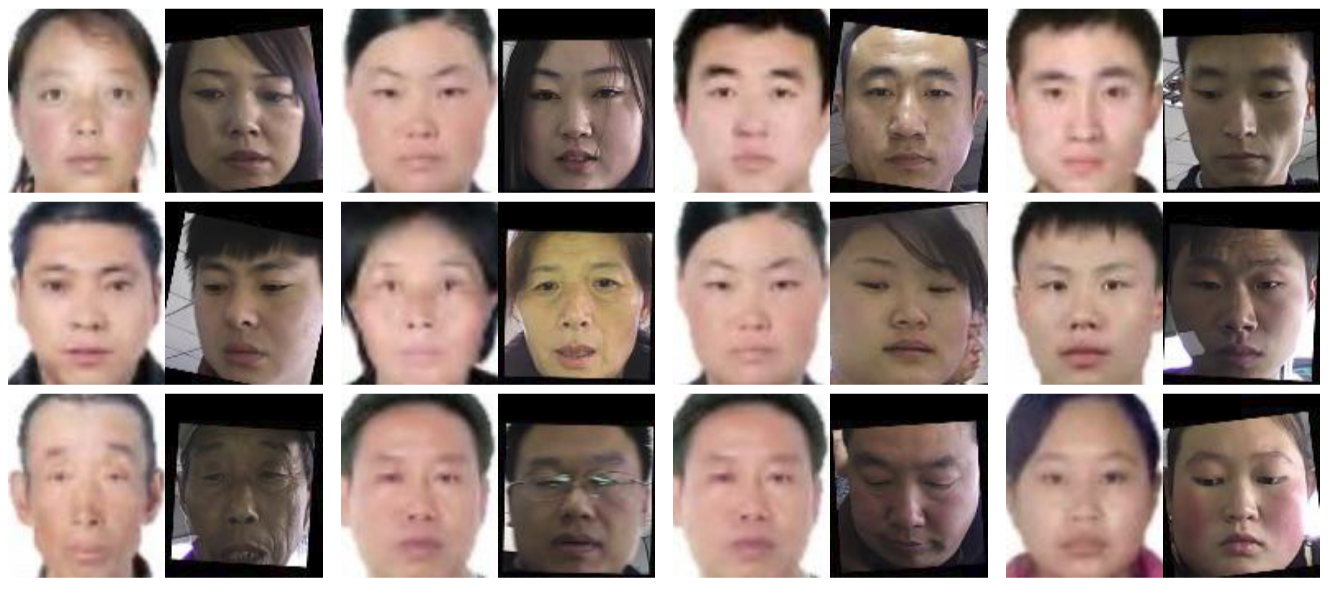}
    \vspace{-2.0em}\caption{False accept pairs}
\end{subfigure}\\
\vspace{0.5em}
\begin{subfigure}[b]{\linewidth}
    \centering
    \includegraphics[width=\linewidth]{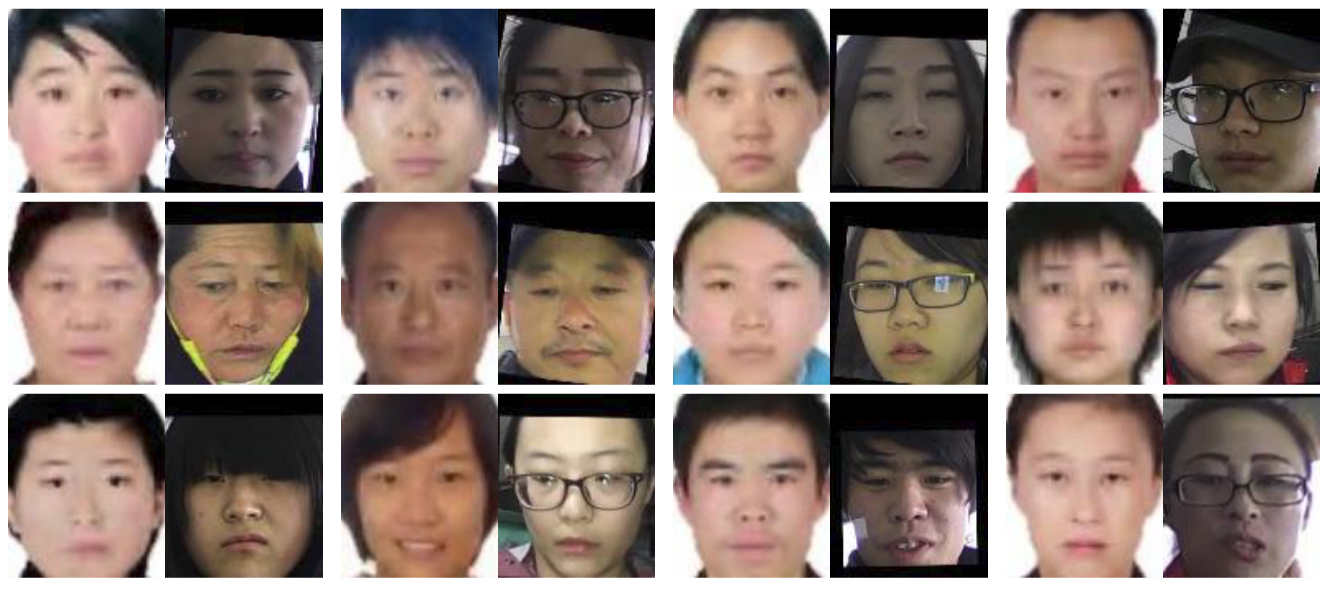}
    \vspace{-2.0em}\caption{False reject paris}
\end{subfigure}
\vspace{-1.5em}
\caption{Example false classified images by our model on ID-Selfie-A dataset at FAR=$0.1\%$.}
\label{fig:false_classification_images}
\vspace{-1.0em}
\end{figure}

The results are shown in Table~\ref{tab:existing_methods}. As one can see, although all the existing methods perform well on the unconstrained face dataset, there is a large performance drop when we test them on the ID-Selfie-A dataset. This is consistent with our observation in Section~\ref{sec:intro} that the characteristics of images and difficulties in the two problems are very different. In comparison, the proposed method significantly improves the performance on the target problem.
Some false accept and false reject image pairs of our model are shown in Figure~\ref{fig:false_classification_images}. From the figure, we can see that most of the selfies in the genuine pairs either have a makeup or their appearance changes drastically because of aging. Besides, many impostor pairs look surprisingly similar, and because of low quality of the ID image, it is hard to find fine-grained clues to tell they are actually different people.

\subsection{Effect of Dataset Size}
\label{sec:exp_dataset}

\begin{figure}
\center
\includegraphics[width=\linewidth]{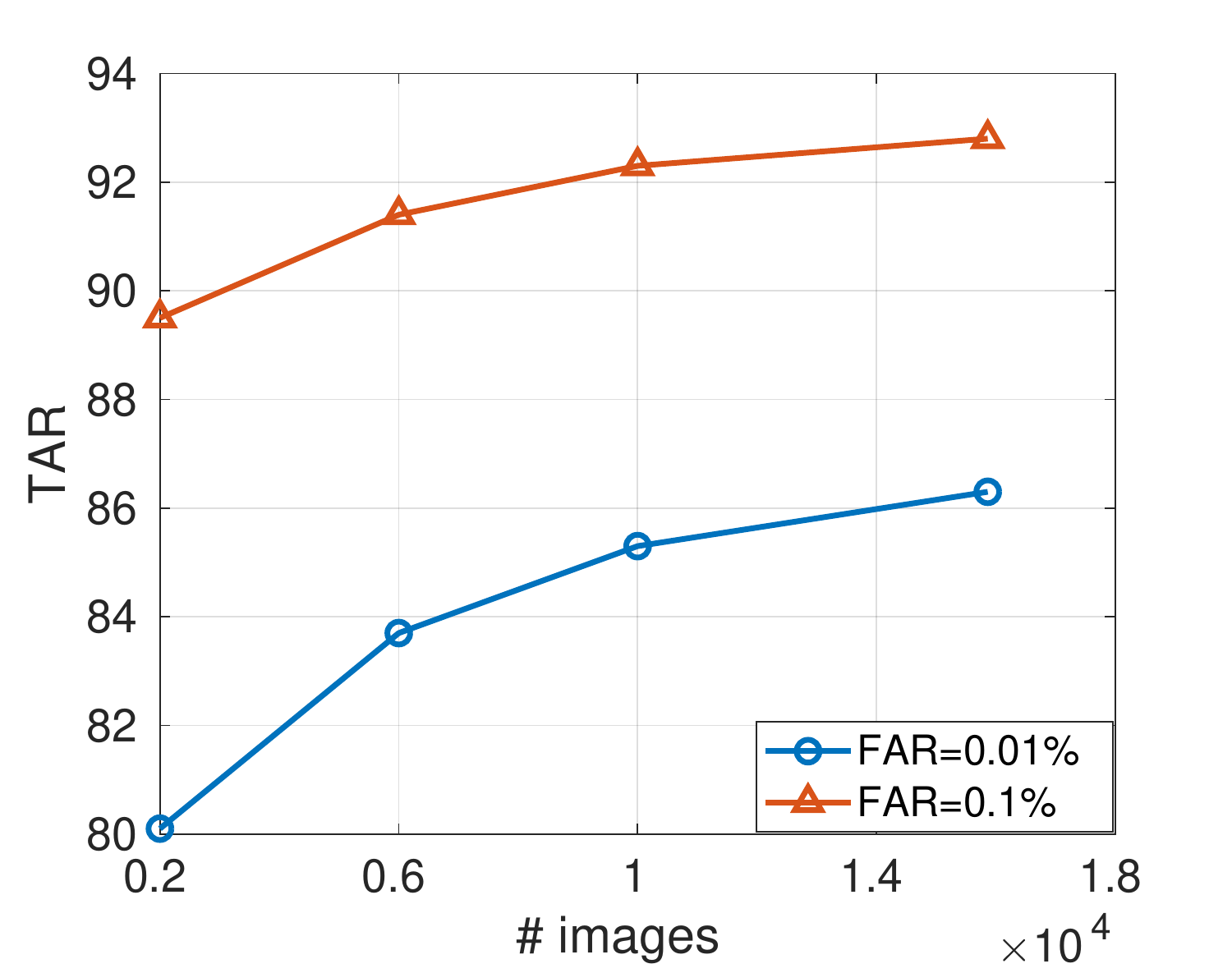}
\caption{Performance when training on subsets of different sizes on ID-Selfie-A dataset. The subsets are randomly selected from the training splits. The performance is reported by taking the average of the five folds.}
\label{fig:training_size}
\end{figure}

In the previous sections, we fix the dataset size and conduct a cross validation to test the performance of different matchers and training strategies. Here we want to explore how much the size of the training dataset could affect our domain-specific network and whether there is a potential for improvement by acquiring more training data. We conduct the same five-fold cross validation, where in each fold we keep the test split unchanged but randomly select a subset of the ID-Selfie pairs in the training splits and report the average performance across the five folds. In particular, we select $1,000$,
$3,000$, $5,000$ and all ($7,932$) image pairs for training. The resulting TAR along with the dataset size is shown in Figure~\ref{fig:training_size}. For both FAR=$0.01\%$ and FAR=$0.1\%$, the performance keeps increasing as the training dataset becomes larger. Notice that we are increasing the size of dataset linearly, which means the relative growth rate of the dataset size is decreasing, yet we can still observe a trend of increasing performance for larger datasets. More performance gain could be expected if we can increase the size of the dataset by one or two orders of magnitude.

\subsection{Cross-dataset Performance Evaluation}
\label{sec:exp_crosseval}
Although it is an important application to match an ID document photos to selfies from stationary cameras, in many other scenarios, the selfies could be captured by different devices including different mobile phones. An ideal model is supposed to perform robustly in all these cases. Therefore, we train a model on the entire ID-Selfie-A dataset and test it on ID-Selfie-B dataset, whose selfies are from different sources. In testing, for subjects in ID-Selfie-B dataset that have more than one selfie images, we fuse their feature vectors by taking the average vector. The results are shown in Table~\ref{tab:performance_dataset_b}. For comparison, we also show the performance of the existing methods. Our model performs the best on ID-Selfie-B, also higher than the base model which has not been fine-tuned on ID-Selfie-A dataset. This indicates that the face representation learned from the ID-Selfie-A dataset is not only discriminative for ID vs. stationary camera pairs, but also useful for other ID-selfie datasets. It also suggests that training on a mixed dataset of images from different sources could be helpful for the performance on all the sub-problems.

\begin{table}[t]
\scriptsize
\begin{center}
\begin{tabularx}{\linewidth}{Xccc}
\toprule
Method                                       & VR(\%)@FAR$=0.01\%$      & VR(\%)@FAR$=0.1\%$  \\
\midrule
COTS                                         & $13.97$                  & $30.91$           \\
CenterFace~\cite{wen2016discriminative}      & $17.69$                  & $35.20$            \\
SphereFace~\cite{liu2017sphereface}          & $34.82$                  & $54.19$               \\ \hline
\textit{Base model}                        & $70.87$                  & $86.77$             \\
\textit{DocFace}                   & $\mathbf{78.40}$         & $\mathbf{90.32}$      \\
\bottomrule
\end{tabularx}
\end{center}
\vspace{-1.5em}
\caption{\small Cross-dataset evaluation on the ID-Selfie-B dataset. The ``base model" is only trained on MS-Celeb-1M. The model \textit{DocFace} has been fine-tuned on ID-Selfie-A. Our models are shown in italic style.}
\label{tab:performance_dataset_b}
\vspace{-1.5em}
\end{table}

\section{Conclusion}
In this paper, we propose a new method, DocFace, which uses transfer learning techniques with a new loss function, Max-margin Pairwise Score (MPS) loss, to fine-tune a pair of sibling networks for the ID document photo matching problem. By using two private datasets, we evaluate the performance of the DocFace and existing unconstrained face matchers on the ID document matching problem. Experiment results show that general face matchers perform poorly on this problem because it involves many different difficulties and DocFace significantly improves the performance of current face matchers on this problem. We also show that testing performance increases steadily with the size of the training set, which implies that additional training data could lead to better recognition performance.

{\small
\bibliographystyle{ieee}
\bibliography{egbib}
}

\end{document}